\documentclass[sigconf]{acmart}
\AtBeginDocument{%
  }

\setcopyright{acmlicensed}
\copyrightyear{2025}
\acmYear{2025}
\acmDOI{XXXXXXX.XXXXXXX}
\acmISBN{978-1-4503-XXXX-X/2018/06}
\acmConference[]{}{}{}
\usepackage[nohyperlinks, printonlyused, nolist]{acronym}
\usepackage{multirow}
\usepackage{siunitx}
\usepackage{pifont}
\usepackage{cleveref}
\begin{document}

\title{Using Sign Language Production as Data Augmentation to enhance Sign Language Translation}
\author{Harry Walsh}
\email{harry.walsh@surrey.ac.uk}
\orcid{0000-0003-4003-7847}
\affiliation{%
  \institution{University of Surrey}
  \city{Guildford}
  \country{United Kingdom}
}

\author{Maksym Ivashechkin}
\email{m.ivashechkin@surrey.ac.uk}
\affiliation{%
\institution{University of Surrey}
  \city{Guildford}
  \country{United Kingdom}
}

\author{Richard Bowden}
\email{r.bowden@surrey.ac.uk}
\affiliation{%
\institution{University of Surrey}
  \city{Guildford}
  \country{United Kingdom}
}


\begin{abstract}
\label{sec:abstract}

\noindent
Machine learning models fundamentally rely on large quantities of high-quality data. Collecting the necessary data for these models can be challenging due to cost, scarcity, and privacy restrictions. Signed languages are visual languages used by the deaf community and are considered low-resource languages. Sign language datasets are often orders of magnitude smaller than their spoken language counterparts. \acf{slp} is the task of generating sign language videos from spoken language sentences, while \acf{slt} is the reverse translation task. Here, we propose leveraging recent advancements in \ac{slp} to augment existing sign language datasets and enhance the performance of \ac{slt} models. For this, we utilize three techniques: a skeleton-based approach to production, sign stitching, and two photo-realistic generative models, SignGAN and SignSplat. We evaluate the effectiveness of these techniques in enhancing the performance of \ac{slt} models by generating variation in the signer's appearance and the motion of the skeletal data. Our results demonstrate that the proposed methods can effectively augment existing datasets and enhance the performance of \ac{slt} models by up to 19\%, paving the way for more robust and accurate \ac{slt} systems, even in resource-constrained environments.
\\
\end{abstract}



%
%
%
\markboth{Nomenclature}{Nomenclature}
\begin{acronym}[iccv] 

\acro{bmvc}[BMVC]{British Machine Vision Conference}
\acro{iccv}[ICCV]{International Conference on Computer Vision}

\acro{ai}[AI]{Artificial Intelligence}
\acro{ar}[AR]{Augmented Reality}
\acro{sdk}[SDK]{Software Development Kit}

\acrodefplural{rnn}[RNNs]{Recurrent Neural Networks}
\acrodefplural{cnn}[CNNs]{Convolutional Neural Networks}
\acrodefplural{hmm}[HMMs]{Hidden Markov Models}
\acrodefplural{gru}[GRUs]{Gated Recurrent Units}
\acrodefplural{crf}[CRFs]{Conditional Random Fields}
\acrodefplural{gan}[GANs]{Generative Adversarial Networks}
\acrodefplural{gpu}[GPUs]{Graphic Processing Units}

\acrodefplural{mdn}[MDNs]{Mixture Density Networks}

\acro{asl}[ASL]{American Sign Language}
\acro{aussl}[Auslan]{Australian Sign Language}
\acro{btg}[BTG]{Bracketing Transduction Grammar}
\acro{bpe}[BPE]{Byte Pair Encoding}
\acro{bsl}[BSL]{British Sign Language}
\acro{bleu}[BLEU]{Bilingual Evaluation Understudy}
\acro{bobsl}[BOBSL]{BBC-Oxford British Sign Language}
\acro{blstm}[BLSTM]{Bidirectional Long Short-Term Memory}
\acro{bslcpt}[BSLCP\textbf{T}]{BSL Corpus \textbf{T}}
\acro{cnn}[CNN]{Convolutional Neural Network}
\acro{crf}[CRF]{Conditional Random Field}
\acro{cslr}[CSLR]{Continuous Sign Language Recognition}
\acro{ctc}[CTC]{Connectionist Temporal Classification}
\acro{c4a}[C4A]{Content4All}
\acro{dl}[DL]{Deep Learning}
\acro{dgs}[DGS]{German Sign Language - Deutsche Gebärdensprache}
\acro{dsgs}[DSGS]{Swiss German Sign Language - Deutschschweizer Geb\"ardensprache}
\acro{dtw}[DTW]{Dynamic Time Warping}
\acro{dtwmje}[DTW-MJE]{Dynamic Time Warping Mean Joint Error}
\acro{fc}[FC]{Fully Connected}
\acro{ff}[FF]{Feed Forward}
\acro{fps}[fps]{frames per second}
\acro{gan}[GAN]{Generative Adversarial Network}
\acro{gpu}[GPU]{Graphics Processing Unit}
\acro{gru}[GRU]{Gated Recurrent Unit}
\acro{gtpt}[G2PT]{Gloss-to-Pose Transformer}
\acro{gtp}[G2P]{Gloss-to-Pose}
\acro{gts}[G2S]{Gloss-to-Sign}
\acro{gtt}[G2T]{Gloss-to-Text}
\acro{gs}[GS]{Gloss Selection}
\acro{gr}[GR]{Gloss Reordering}
\acro{gt}[GT]{ground truth}
\acro{hmm}[HMM]{Hidden Markov Model}
\acro{hpe}[HPE]{Hand Pose Enhancer}
\acro{hoh}[HOH]{Hard of Hearing}
\acro{hns}[HamNoSys]{Hamburg Notation System}

\acro{isl}[ISL]{Irish Sign Language}
\acro{lstm}[LSTM]{Long Short-Term Memory}
\acro{lsf}[LSF]{French Sign Language}
\acro{llms}[LLMs]{Large Language Models}
\acro{mha}[MHA]{Multi-Headed Attention}
\acro{mo}[MoCap]{Motion Capture}
\acro{mtc}[MTC]{Monocular Total Capture}
\acro{mse}[MSE]{Mean Squared Error}
\acro{mdn}[MDN]{Mixture Density Network}
\acro{mdgs}[MeineDGS]{Meine DGS Annotated}
\acro{mdgst}[mDGS\textbf{T}]{meineDGS\textbf{T}}
\acro{mdgsth}[mDGS\textbf{T}-\textbf{H}]{meineDGS\textbf{T}-\textbf{HARD}}
\acro{mdgste}[mDGS\textbf{T}-\textbf{E}]{meineDGS\textbf{T}-\textbf{EASY}}
\acro{mt}[MT]{Machine Translation}

\acro{nmt}[NMT]{Neural Machine Translation}
\acro{nlp}[NLP]{Natural Language Processing}
\acro{nar}[NAR]{Non-AutoRegressive}
\acro{nsvq}[NSVQ]{Noise Substitution Vector Quantization}
\acro{ph12}[PHOENIX12]{RWTH-PHOENIX-Weather-2012}
\acro{ph14}[PHOENIX14]{RWTH-PHOENIX-Weather-2014}
\acro{ph14t}[PHOENIX14\textbf{T}]{RWTH-PHOENIX-Weather-2014\textbf{T}}
\acro{pof}[POF]{Part Orientation Field}
\acro{pos}[POS]{Part Of Speech}
\acro{pt}[PT]{Progressive Transformer}
\acro{paf}[PAF]{Part Affinity Field}
\acro{pttt}[P2TT]{Pose-to-Text Transformer}
\acro{ptt}[P2T]{Pose-to-Text}
\acro{pts}[P2S]{Pose-to-Sign}
\acro{ptgtt}[P2G2T]{Pose-to-Gloss-to-Text}
\acro{pca}[PCA]{Principal Component Analysis}
\acro{relu}[RELU]{Rectified Linear Units}
\acro{rnn}[RNN]{Recurrent Neural Network}
\acro{rouge}[ROUGE]{Recall-Oriented Understudy for Gisting Evaluation}
\acro{vqvae}[VQ-VAE]{Vector Quantized Variational Autoencoders}
\acro{vq}[VQ]{Vector Quantisation}
\acro{vae}[VAE]{Variational Autoencoders}
\acro{vtt}[V2T]{Video-to-Text}
\acro{vlp}[VLP]{Visual Language Pre-training}
\acro{sgd}[SGD]{Stochastic Gradient Descent}
\acro{sla}[SLA]{Sign Language Assessment}
\acro{slr}[SLR]{Sign Language Recognition}
\acro{slt}[SLT]{Sign Language Translation}
\acro{slp}[SLP]{Sign Language Production}
\acro{smt}[SMT]{Statistical Machine Translation}
\acro{slo}[SLO]{Spoken Language Order}
\acro{so}[SO]{Sign Language Order}
\acro{snr}[S\&R]{Select and Reorder}
\acro{sse}[SSE]{Sign Supported English}
\acro{stt}[S2T]{Sign-to-Text}
\acro{stgtt}[S2G2T]{Sign-to-Gloss-to-Text}
\acro{sio}[SIO]{Sign Language Order}
\acro{spo}[SPO]{Spoken Language Order}
\acro{ttgt}[T2GT]{Text-to-Gloss Transformer}
\acro{ttpt}[T2PT]{Text-to-Pose Transformer}
\acro{ttp}[T2P]{Text-to-Pose}
\acro{ttg}[T2G]{Text-to-Gloss}
\acro{ttg++}[T2G++]{Text-to-Gloss++}
\acro{tth}[T2H]{Text-to-HamNoSys}
\acro{ttgth}[T2G2H]{Text-to-Gloss-to-HamNoSys}
\acro{ttgtp}[T2G2P]{Text-to-Gloss-to-Pose}
\acro{tts}[T2S]{Text-to-Sign}
\acro{ttsse}[T2SSE]{Text to Sign Supported English}
\acro{ttspog}[T2SPOG]{Text to Spoken Language Order Gloss}
\acro{wer}[WER]{Word Error Rate}
\acro{wmt14}[WMT2014]{WMT2014 German-English}

\end{acronym}
\maketitle

\section{Introduction}
\label{sec:intro}
Sign languages are visual languages that use multiple articulators, such as the hands, body, and facial expressions to convey meaning. They are natural languages used by the Deaf community and possess their own grammar and syntax \cite{stokoe1980sign}. Sign Languages can be classified as low-resource languages. Given their visual nature, recording and annotating them poses a significant challenge, especially in the quantities comparable to the spoken language domain. Acquiring real-world data presents numerous challenges, such as high collection and labelling costs, scarcity, and privacy restrictions \cite{bragg2019sign}. Given that machine learning models rely on large quantities of high-quality data for training, this has been a limiting factor in computational Sign language research.


Sign language datasets are orders of magnitude smaller than their spoken language counterparts \cite{bragg2019sign}. For instance, a common dataset used as a baseline in the field, the \ac{ph14t} dataset, only contains approximately 9,000 sentences \cite{camgoz2018neural}. While larger datasets exist, the linguistic annotation, which is required for state-of-the-art approaches, is often missing, incomplete, or non-existent \cite{albanie2021bbc}. Primarily due to the costs and expertise required to create such annotations.



A prevalent solution to address data scarcity is to augment existing datasets. Data augmentation encompasses a variety of techniques designed to artificially increase the size and diversity of training samples without collecting new data \cite{feng2021survey}. In contrast, an alternative approach is to generate synthetic data. Synthetic data is defined as artificially generated information that mimics the statistical properties and patterns of real-world data, produced via computational methods such as algorithms, simulations, or increasingly sophisticated generative AI models. Data augmentation and generation have become widely adopted practices in various domains, such as \ac{nlp} \cite{shorten2021text}.

\begin{figure}
    \centering
    \includegraphics[width=0.95\linewidth]{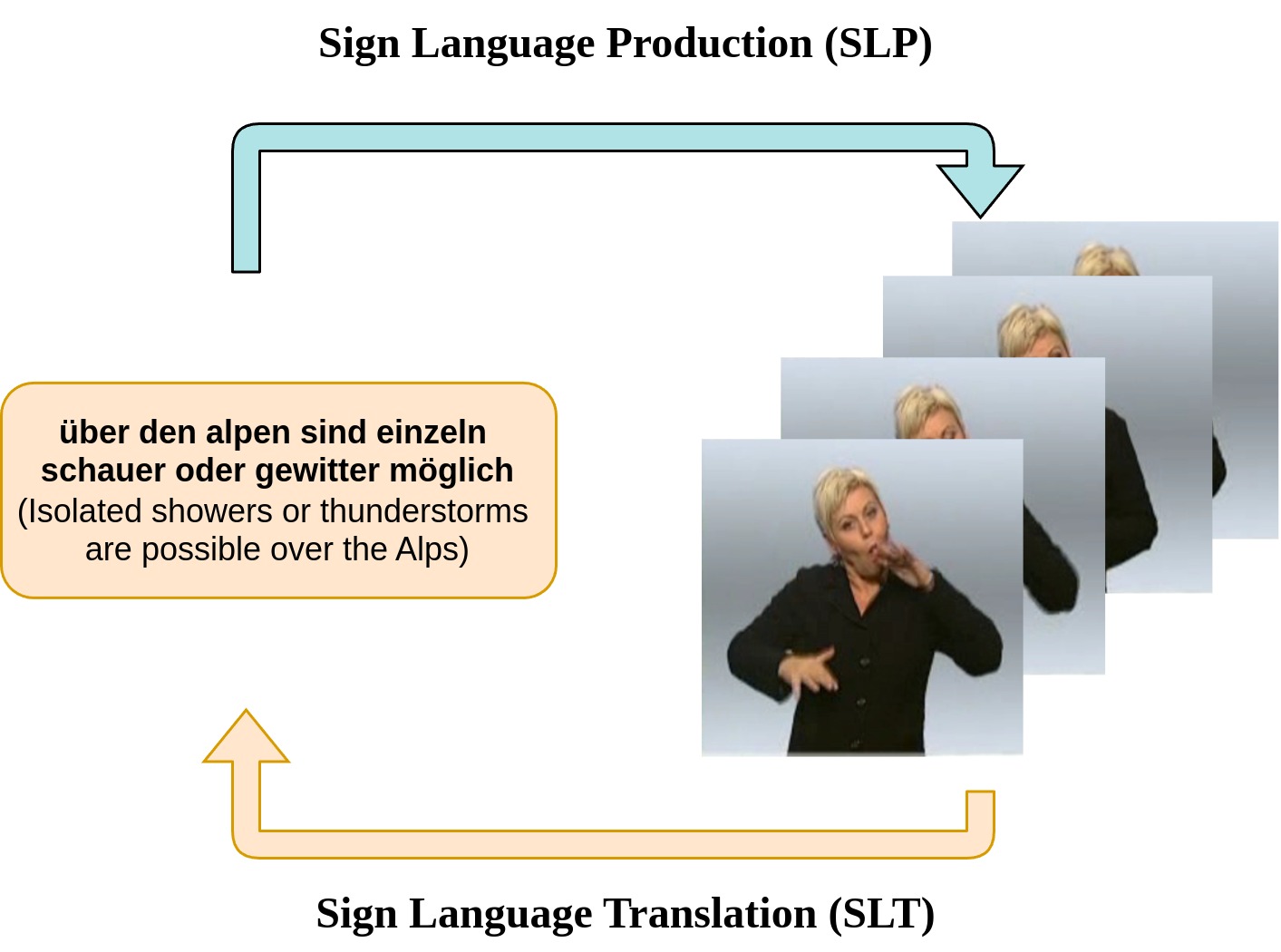}
    \caption{A visual overview of the \ac{slt} and \ac{slp} tasks. The \ac{slt} task is the process of translating sign language video into spoken language text. The \ac{slp} task is the reverse, generating sign language video from spoken language text.}
    \label{fig:slt_slp_overview}
\end{figure}

As illustrated in \Cref{fig:slt_slp_overview}, \ac{slt} is the task of predicting a spoken language translation from a sign language video, \ac{stt}. Whereas, \ac{slp} is the reverse task, aiming to generate sign language videos given spoken language sentences, \ac{tts}. In this paper, we leverage three \ac{slp} techniques to augment and generate synthetic data \cite{ivashechkin2025signsplat, walsh2024sign, saunders2020everybody}. We then leverage this data to enhance the translation ability of \ac{slt} models \cite{camgoz2018neural, zhou2023gloss}. We note the limitations of these approaches, as they often lack non-manual features, and significantly more work is required for the approaches to fully capture the subtleties of the language. However, we suggest that they capture enough of the manual features to be a useful tool for pre-training or to supplement data during training. 



\begin{figure*}
    \centering
    \includegraphics[width=\linewidth, trim=50 50 50 85, clip]{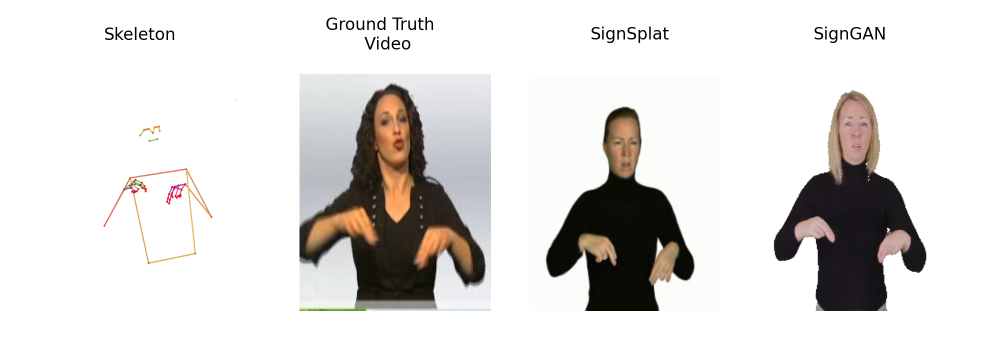}
    \caption{An example from the \ac{ph14t} dataset, showing left to right: skeleton pose, original video, SignSplat Avatar, and SignGAN Avatar.}
    \label{fig:rep_plot}
\end{figure*}

Some \ac{slt} approaches perform translation solely from video \cite{wong2024sign2gpt, zhou2023gloss}, \ac{vtt}, and others leverage a skeleton pose representation \cite{tang2021graph, chen2022two, fish2025geosign}, \ac{ptt}. Our first technique focuses on enhancing \ac{ptt} architectures and therefore synthesises synthetic skeleton data. For this, we leverage sign stitching \cite{walsh2024sign}, an approach that uses a dictionary of pre-recorded isolated signs and joins them together to create continuous sign language sequences. Previous research has suggested many problems with alternative methods that attempt to directly regress a sequence of poses from the spoken language text \cite{saunders2021continuous}. Mainly regression to the mean, which causes under-articulated and often incomprehensible signing. For this reason, we choose a stitching approach as it is guaranteed to produce expressive natural sequences.

 The next two techniques focus on the \ac{vtt} architectures for which both a \ac{gan} \cite{saunders2021signing} and Gaussian Splatting \cite{ivashechkin2025signsplat} approaches are tested. Both methods use skeleton pose to generate a photo-realistic signer.  \ac{gan}s often struggle to generalise to out-of-domain poses and suffer from noisy conditioning.  This can lead to several artefacts in production. Whereas, a Gaussian Splatting-based approach that uses SMPL-X mesh in addition to hammer features provides additional constraints on the generation process, resulting in fewer artefacts. Previous works emphasised that the quality of the pre-training data is paramount in achieving strong performance in downstream tasks \cite{liu2024best, firdoussi2024maximizing} and low-quality data can even harm model performance \cite{shumailov2023curse, seddik2024bad}.

\noindent
We define the contributions of this paper as follows;
\begin{enumerate}
    \item To the best of our knowledge, we are the first to propose generating synthetic skeleton sequences for \ac{slt}.
    \item We propose augmenting the appearance of sign language data, testing two photo-realistic avatar models, SignGAN \cite{saunders2021signing} and SignSplat \cite{ivashechkin2025signsplat}.
    \item We evaluate the effectiveness of these techniques in enhancing the performance of \ac{slt} models, demonstrating that they can significantly improve translation accuracy.
\end{enumerate}

 The rest of this paper is organized as follows. Section \ref{sec:related_work} reviews related works in the field of data augmentation, generation, \ac{slp}, and \ac{slt}. Section \ref{sec:methodology} elaborates on the methodology for augmenting the data and the translation models used to verify its effectiveness. Section \ref{sec:experiments} details the evaluation protocol before moving on to the experimental results. Finally, Section \ref{sec:conclusion} concludes the paper and suggests directions for future work.

\section{Related Work}
\label{sec:related_work}

Data augmentation and generation have enhanced machine learning models. We review approaches from other fields that leverage a range of techniques to improve system performance. Before moving on to the field of \ac{slp}, which we leverage to produce synthetic skeleton data and augment the appearance of the signers. Finally, we review the \ac{slt} models that we use to evaluate the effectiveness of the proposed techniques.

\subsection{Data Augmentation and Generation}


 Synthetic data is commonplace where the rules and constraints are well understood \cite{deng2025graspvla, guo2022learning, erez2015simulation}. This can also help deal with rare scenarios where capturing additional data is expensive or unobtainable due to privacy concerns. In the related field of action recognition, simulated data was shown to be very effective and helped boost the recognition performance when fine-tuned on real data \cite{kim2022transferable}. Statistical methods have been used to generate data by learning the underlying distribution of the real data. These methods often involve techniques such as Monte Carlo Simulation, Hidden Markov Models (HMMs), and other probabilistic models \cite{hao2024synthetic}. However, more recently, they have been replaced with deep learning based approaches that have shown superior performance.

 Deep learning based approaches have utilised a variety of models in order to augment and generate data, such as \ac{gan} \cite{goodfellow2020generative}, \ac{vae} \cite{kingma2013auto}, Diffusion \cite{ho2020denoising}, and transformer-based models \cite{vaswani2017attention}. All of which have been shown to generate high-quality synthetic data \cite{hao2024synthetic}.  \ac{gan}s and diffusion models have been used to generate high-quality images and videos for several applications like medical imaging \cite{dunn2019deepsynth}, object detection \cite{niemeyer2021giraffe}, computer vision \cite{ho2022video}, etc. \ac{llms} are transformer-base models, that are commonly used to generate synthetic data for various tasks such as knowledge distillation \cite{gou2021knowledge, hinton2015distilling}, mathematical reasoning \cite{luo2023wizardmath, yu2023metamath}, coding \cite{hu2023instructcoder}, or even refining its own predictions. For instance, LLAMA 3.1 is fine-tuned on synthetic data to tailor the output to a desired style \cite{grattafiori2024llama}.

 Given that sign language is classified as a low-resource language, authors have previously attempted to augment monolingual text data to generate synthetic gloss\footnote{Gloss is the written word associated with each sign performed in a sequence.} sequences. Gloss is a commonly used text intermediary for \ac{slp} and \ac{slt}, but is costly and time-consuming to create, and has been noted as a limiting factor in expanding state-of-the-art approaches to larger domains of discourse. Moryossef et al. \cite{moryossef2021data} focused on \ac{gtt} translation, while Yao et al. \cite{yao2024semi} focused on the reverse, \ac{ttg}. Both approaches used a deep translation model and a rule-based system to generate gloss sequences from spoken language sentences. Recently, Abdullah et al. leveraged  GPT-4o, a \ac{llms} to generate gloss sequences from spoken language sentences. Using this data, they showed improved performance \cite{abdullah2025state}. The approaches were able to show the benefits of utilising synthetic data. 
 As the field advances toward end-to-end methods, which do not require linguistic annotations, the relevance of such approaches has been reduced. 
 Therefore, we propose creating synthetic skeleton sequences and appearance augmentation for enhanced \ac{slt}.

When augmenting data, there are two main approaches to integrating it into a model. Some methods suggest first pre-training on the augmented data, followed by a second stage of fine-tuning on real data at a lower learning rate. Others propose simultaneously training on both real and augmented data \cite{feng2021survey}. In this work, we experiment with both of these approaches.

\subsection{Sign Language Production}

\ac{slp} is the task of generating realistic sign language sequences from spoken language text. To accomplish this, several intermediate representations can be employed within the translation pipeline. A common approach first generates a gloss representation, followed by a skeleton pose production that is used to drive a photo-realistic signer \cite{stoll2018sign, 10.1007/s11263-019-01281-2, walsh2024sign, saunders2022signing}. An example of the skeleton pose representation can be seen in \Cref{fig:rep_plot}

Early approaches to \ac{slp} used a graphical avatar-driven systems\cite{bangham2000virtual, cox2002tessa, efthimiou2012dicta, elghoul2011websign, zwitserlood2004synthetic}. Some of the approaches looked for legal phrases, and, thus are limited to pre-recorded phrases \cite{cox2002tessa}. An alternative approach performed a translation to a linguistic notation, then opted to play each sign in sequence with unnatural transitions in between \cite{bangham2000virtual}. The avatar was often unrealistic; and as a result, these early approaches were unpopular with the deaf community \cite{kipp2011assessing}.

 Deep learning based approaches have improved the realism of the signing sequences. Initially being tackled with \acp{rnn} \cite{stoll2018sign, 9093516}, later being improved upon using transformer based architectures \cite{saunders2021continuous, saunders2020adversarial, saunders2020progressive}.  However, models that attempt to directly regress skeleton pose often suffer from regression to the mean. This is caused by the model attempting to minimise their loss function and therefore, they result in under-articulated and incompressible signing. Given that these models can also hallucinate, we avoid them for the \ac{ptt} task.

 Alternatively, other methods used a combination of deep learning and a pre-recorded dictionary of isolated signs.  These methods attempt to learn the co-articulation between isolated signs in order to create fully continuous natural sequences. Walsh et al. \cite{walsh2024sign} used a 7-step pipeline that included non-manual features. By including features such as the timing as well as the frequency components of a sequence, the method is able to capture signed prosody\footnote{the natural rhythm, stress and intonation used to convey additional meaning in sign language}, this is utilized in \Cref{sec:p2t} to generate our synthetic skeleton data.  Other methods instead opt to use a diffusion model \cite{tang2024discrete} or a transformer \cite{xu2021simple} to predict the transitions between isolated signs.  Saunders et al. \cite{saunders2022signing} proposed a keyframe selection network to select specific frames from isolated signs, and by concatenating and interpolating between them, were able to produce continuous sequences. The output was used to drive the SignGAN model, a model capable of generating photo-realistic sign language video given skeleton pose and a style image. The model is trained through a min-max game performed by the generator and discriminator.  Here, the SignGAN model is used to generate appearance variations for the \ac{vtt} task in \Cref{sec:v2t}.

Taking inspiration from computer graphics and the field of 3D rendering, Ivashechkin et al. \cite{ivashechkin2025signsplat} proposed a Gaussian Splatting based approach to generate photorealistic sign language video by attaching 3D Gaussian splatting primitives to the SMPL-X~\cite{SMPL-X:2019} human mesh model.
While alternative Gaussian splatting approaches~\cite{hu2024expressive,moon2024exavatar} also tackle the problem of expressive avatar rendering, the \emph{SignSplat} approach adds regularization and constraints to the underlying mesh geometry and appearance to minimize rendering artifacts and enforce the physical limits of a human.
Furthermore, proposing a sign-stitching mechanism for gloss interpolation. Once again, we leverage this method in \Cref{sec:v2t}. Both a SignGAN and a SignSplat avatar are shown on the right side of \Cref{fig:rep_plot}, respectively.

\subsection{Sign Language Translation}

 Initially, the field focused on isolated \ac{slr}, which aims to produce the corresponding gloss for a short video containing a sign \cite{imashev2020k, joze2018ms, li2020word}.  Later, the field progressed to the more challenging task of Continuous \ac{slr}, which requires multiple signs to be recognised within a sequence \cite{hao2021self, hu2022temporal, koller2019weakly, min2021visual}.  \ac{slt} requires an additional translation to spoken language.

 Similar to the \ac{slp} field, the task was initially tackled using an RNN \cite{camgoz2018neural}, before moving on to a transformer-based architecture \cite{camgoz2020sign} which employed a \ac{ctc} loss on the encoder to simultaneously learn \ac{slr} and \ac{slt} in a single model. Modifying the network to work with skeleton keypoints, has become the standard for evaluating skeleton based \ac{slp} \cite{saunders2021continuous, saunders2020adversarial, saunders2020progressive, walsh2024data, walsh2024sign}. Therefore, we use this architecture in the \ac{ptt} experiments in \Cref{sec:p2t}. Other approaches have made use of both skeleton keypoints and video features in a single model \cite{chen2022two, li2025unisign}. While some rely solely on keypoints as features \cite{tang2021graph, fish2025geosign}.  Skeleton keypoints are appearance agnostic, while generally being computationally inexpensive when compared to video-based models.  However, they can be susceptible to noise.

 Several other video-based approaches have been proposed. Wong et al. were able to leverage the power of \ac{llms} and adapt them for the use of sign language translation \cite{wong2024sign2gpt}. While Zhou et al. \cite{zhou2023gloss} initialised their models using pre-trained visual and language encoders. Using Contrastive Language-Image Pre-Training (CLIP) \cite{DBLP:journals/corr/abs-2103-00020} the approach was able to learn an effective representation that helps in the downstream translation task.  We adopt this architecture when experimenting with our \ac{vtt} approach in \Cref{sec:v2t}.


 It is commonplace for translation models to apply visual augmentations to the training data in order to make them more robust to colour variations and signer placement in the video. Such strategies include colour jittering, random Gaussian noise plus random, cropping, rotation, and scaling \cite{algafri2025sslr, zhou2023gloss, wong2024sign2gpt, wong2025signrep}.  However, to the best of our knowledge, we are the first to augment the appearance entirely to a new signer and use this data for training a \ac{slt} model.

\section{Methodology}
\label{sec:methodology}
First, we explain the \ac{slp} techniques that are used to augment sign language data. We then describe the \ac{slt} models that are used to conduct the translation experiments.

\subsection{Data Augmentation Techniques}
\label{sec:data_augmentation}


The three proposed augmentation strategies can be categorised into two types: First, skeleton pose augmentation, where we use MediaPipe skeletons as a representation for sign language. By generating new synthetic sequences from isolated signs, we can create variations in the lexical form of the signs, in addition to altering the speed and order. Second, photo-realistic augmentations, where we use generative models to produce realistic sign language videos. This approach allows us to augment the appearance of the signer and create more variations in camera angle and background.

\subsubsection{Sign Stitching}

For a given dataset, we construct a dictionary of the performed signs. This dictionary comprises, for each sign, a video of an isolated sign and its corresponding spoken language gloss tag. From each video, we extract a 3D skeleton pose using the method described in \cite{10193629}. This method employs inverse kinematics and a neural network to uplift a 2D MediaPipe pose to 3D. Solving for joint angles allows this method to enforce physiological constraints of the human body. The dictionary is stored as sequences of joint angles, a representation that facilitates the application of a canonical skeleton during pose generation. This ensures consistent bone lengths across all signers, irrespective of the original performer's physique.

Let the dictionary be denoted as \( \mathcal{D} \), where each entry maps a gloss \( y \) to its corresponding sequence of angles \( A \). This can be expressed as:

\[
\mathcal{D} = \{ (y_i, A_i) \mid y_i \in \mathcal{Y}, A_i = (a_{i,1}, a_{i,2}, \ldots, a_{i,U_i}) \},
\]

where \( \mathcal{Y} \) is the set of all unique glosses in the dataset, and \( A_i \) is the sequence with length \( U_i \) for gloss \( y_i \). If a specific sign is not available in \( \mathcal{D}\), we employ a word embedding model to vectorise its gloss tag.
Then we select the most similar sign from the dictionary based on the highest cosine similarity in the embedding space.

\textbf{Sequence Generation} - Given an input sequence of glosses \(Y = (y_{1}, y_{2}, \ldots, y_{G})\), their corresponding durations \(D = (d_{1}, d_{2}, \ldots, d_{G})\), each of length \(G\), and a low-pass cutoff frequency \( \mathcal{C} \) specified per sequence, we generate a continuous sequence of 3D skeleton poses \(P = (p_{1}, p_{2}, \ldots, p_{U})\) with \(U\) frames. The per-gloss durations \(d_i\) in \(D\) can be approximated using the method in \cite{walsh2024sign}. We summarise this process into three steps: 

\textbf{1. Sign Retrieval: } The angular sequences \(A_i\) for each gloss \(y_i\) in \(Y\) are retrieved from \( \mathcal{D} \). These are then converted from their angular representation to 3D skeleton poses by applying the canonical skeleton. Concurrently, each sign's pose sequence is resampled to match its specified duration \(d_i\) from \(D\).

\textbf{2. Stitching: } The resampled isolated sign pose sequences are concatenated. To ensure smooth coarticulation between the end of one sign and the start of the subsequent sign, transitions are generated using linear interpolation. The number of transition frames is determined by the distance between boundary poses, with the constraint that the transitional movement velocity remains consistent with, or bounded by, the velocities at the sign boundaries.

\textbf{3. Motion Filtering: } Finally, the entire stitched sequence of poses \(P\) is processed using a low-pass Butterworth filter \cite{butterworth1930theory} with the predefined cutoff frequency \(\mathcal{C}\). It is observed that natural signing exhibits distinct motion frequency ranges corresponding to sharpness or smoothness. This filtering step aims to remove sharp, unnatural movements not present in the original data, thereby emulating motion characteristics of the original signer.

\textbf{Additional Augmentations} - To introduce further variation into the skeleton data, two additional augmentation methods are implemented; First, gloss order permutation, where we apply random permutations to the input gloss order \(Y\). Second, speed variation, we vary the total number of frames \(U\) in the generated sequence \(P\), which effectively alters its performance speed.



\subsubsection{SignGAN}

Given a pose sequence, \(P = (p_{1}, p_{2}, \ldots, p_{U})\), the model aims to generate the corresponding video of a photorealistic signer, \(V = (v_{1}, v_{2}, \ldots,v_{U})\) with \(U\) frames. The model is trained using a \ac{gan} architecture \cite{goodfellow2014generative} that comprises a generator and discriminator network. The generator takes the pose sequence as input and generates the corresponding video frames. We train the model on a single signer's appearance. Using a \ac{cnn}, we extract features from a style image and fuse them into the generator at multiple layers. While the discriminator evaluates the realism of the generated frames against real video frames. The generator is an encoder-decoder architecture that consists of a series of convolutional layers. The model uses residual connections to preserve spatial information and improve the quality of the generated frames.

\subsubsection{SignSplat}

We train a Gaussian splatting~\cite{kerbl3Dgaussians} model on multi-view capture data. We primarily utilized the framework from~\cite{ivashechkin2025signsplat}, which exploits the SMPL-X~\cite{SMPL-X:2019} human body parameterization with manually updated human constraints, especially for the hand joints (removing redundant degrees of freedom and limiting the angular range to be more realistic).
To exploit the Gaussian splatting human appearance model for 3D reconstruction from a video input, we propose the following pipeline.
First, we process an input image with MMPose~\cite{mmpose2020} to obtain 2D OpenPose keypoints.
Second, we estimate the intrinsic matrix with fixed focal length and principal point based on the image resolution.
We then run inverse kinematics optimization to fit the SMPL-X parameters to the 2D keypoints.
This involves generating an SMPL-X mesh, applying linear blend skinning to obtain a 3D skeleton, projecting it using the estimated intrinsics, and minimizing the reprojection error.

Due to single-view ambiguities, the 3D reconstruction primarily suffers from poor hand reconstruction, because 2D hand keypoints do not carry depth information.
To mitigate this issue, for the \ac{ph14t} dataset, we exploited the HaMeR~\cite{pavlakos2024reconstructing} hand angle parameters reconstruction for our initialization.
The HaMeR hand angles (in MANO~\cite{MANO:SIGGRAPHASIA:2017} format) have a good 3D prior and can be directly transferred to the SMPL-X mesh.
Consequently, in the overall inverse-kinematics optimization, we only fine-tune the hands with a smaller learning rate.    
Such optimization leads to a better correspondence of the 3D mesh to 2D detections and improved hand-to-hand interaction, essential for finger-spelling.
Once the SMPL-X parameters are optimized, we drive our Gaussian Splatting model to render the signer in real time.

\subsection{\acf{slt} Models}
Both \ac{vtt} and \ac{ptt} \ac{slt} models are transformer-based encoder-decoder architectures. We utilise Sign Language Transformers \cite{camgoz2020sign} for the \ac{ptt} task, and GF-SLT \cite{zhou2023gloss} for the \ac{vtt} task. We detail the difference in the models below.

\subsubsection{Sign Language Transformers}

This approach is commonly used in \ac{slp} work \cite{walsh2024data, walsh2024sign, saunders2021signing, huang2021towards, saunders2020adversarial, saunders2021continuous, saunders2020progressive}, hence we use it for the \ac{ptt} task. The encoder processes the input skeleton pose sequence, \(P\), and is supervised using a \ac{ctc} loss
 so that it performs \ac{slr}.  The decoder autoregressively predicts the spoken language translation, \(X = (x_{1},x_{2}, \ldots,x_{W})\) with \(W\) words. Therefore, the model learns the conditional probability, \(p(X|P)\).

\subsubsection{GF-SLT}

For the \ac{vtt} we use the publicly available GF-SLT architecture. Given that this model comes with a pre-trained vision encoder and language encoder, we use this architecture for computational efficiency. Once again, the model follows the same transformer encoder-decoder architecture.  The encoder processes a sequence of Video frames, \(V\), and the decoder autoregressively predicts the spoken language translation, \(X\). Therefore, the model learns the conditional probability, \(p(X|V)\).  The decoder's embedding layers are initialised from a pre-trained multilingual BART model for improved embeddings.

 Furthermore, this model employs \ac{vlp}, which uses a contrastive learning framework, to learn the alignment between video and text features.  This creates a shared embedding space between the two encoders, which is shown to improve the performance in the downstream translation task.
\section{Experiments}
\label{sec:experiments}

\begin{table*}[htbp!]
\centering
\caption[Sign Stitching Data Augmentation Results]{The results of using sign stitching for data augmentation on the RWTH-PHOENIX-Weather-2014\textbf{T} dataset.}

\small a) Baseline approach and different training strategies. The original data used in training is labelled as ``GT'' (ground truth).
\vspace{1.0em}

\resizebox{0.95\linewidth}{!}{%
\begin{tabular}{r|rrrrr|rrrrr}
\toprule
\multicolumn{1}{l}{}             & \multicolumn{5}{c}{TEST SET}                                                                                                                   & \multicolumn{5}{c}{DEV SET}                                                                                                                    \\
\multicolumn{1}{c|}{Training Data:} & \multicolumn{1}{c}{BLEU-1} & \multicolumn{1}{c}{BLEU-2} & \multicolumn{1}{c}{BLEU-3} & \multicolumn{1}{c}{BLEU-4} & \multicolumn{1}{c|}{ROUGE} & \multicolumn{1}{c}{BLEU-1} & \multicolumn{1}{c}{BLEU-2} & \multicolumn{1}{c}{BLEU-3} & \multicolumn{1}{c}{BLEU-4} & \multicolumn{1}{c}{ROUGE}  \\
\midrule
GT                               & 32.41             & 20.19             & 14.41             & 11.32             & 32.96             & 32.36             & 20.02             & 14.25                     & 11.13             & 33.46             \\
Stitched                         & 16.74                      & 6.88                       & 4.53                       & 3.49                       & 17.72                      & 16.78                      & 7.67                       & 5.29                       & 4.21                       & 18.34                      \\
GT + Stitched                    & 31.84                      & 19.45                      & 13.71                      & 10.73                      & 32.04                      & 31.19                      & 19.63                      & 14.26                      & 11.37                      & 32.57                      \\
GT + Stitched Pre-Training                    & \textbf{37.86}             & \textbf{24.36}             & \textbf{17.58}                      & \textbf{13.68}                      & \textbf{37.61}                      & \textbf{37.26}             & \textbf{23.71}                      & \textbf{17.13}                      & \textbf{13.52}                      & \textbf{37.95}                      \\
    \bottomrule
\end{tabular}
}

\vspace{1.0em}
b) Stitched data pre-training with N random permutations in sign order.
\vspace{1.0em}

\resizebox{0.95\linewidth}{!}{%
\begin{tabular}{r|rrrrr|rrrrr}
\toprule
\multicolumn{1}{l}{}             & \multicolumn{5}{c}{TEST SET}                                                                                                                   & \multicolumn{5}{c}{DEV SET}                                                                                                                    \\
\multicolumn{1}{c|}{Permutations:} & \multicolumn{1}{c}{BLEU-1} & \multicolumn{1}{c}{BLEU-2} & \multicolumn{1}{c}{BLEU-3} & \multicolumn{1}{c}{BLEU-4} & \multicolumn{1}{c|}{ROUGE} & \multicolumn{1}{c}{BLEU-1} & \multicolumn{1}{c}{BLEU-2} & \multicolumn{1}{c}{BLEU-3} & \multicolumn{1}{c}{BLEU-4} & \multicolumn{1}{c}{ROUGE}  \\
\midrule
0                  & \textbf{37.86}             & \textbf{24.36}             & 17.58                      & 13.68                      & 37.61                      & \textbf{37.26}             & 23.71                      & 17.13                      & 13.52                      & 37.95                      \\
1            & 36.22                      & 22.84                      & 16.22                      & 12.65                      & 36.09                      & 36.26                      & 22.98                      & 16.45                      & 12.76                      & 36.49                      \\
3            & 37.04                      & 24.22                      & \textbf{17.69}             & \textbf{13.89}             & \textbf{37.83}             & 37.18                      & \textbf{24.07}             & \textbf{17.57}             & \textbf{13.81}             & \textbf{38.00}             \\
10           & 37.67                      & 24.18                      & 17.23                      & 13.24                      & 37.36                      & 36.47                      & 23.26                      & 16.83                      & 13.16                      & 36.91                      \\
    \bottomrule
\end{tabular}
}

\vspace{1.0em}
c) Stitched data pre-training with variations in the duration of each sequence.
\vspace{1.0em}

\resizebox{0.95\linewidth}{!}{%
\begin{tabular}{r|rrrrr|rrrrr}
\toprule
\multicolumn{1}{l}{}             & \multicolumn{5}{c}{TEST SET}                                                                                                                   & \multicolumn{5}{c}{DEV SET}                                                                                                                    \\
\multicolumn{1}{c|}{Duration scale:} & \multicolumn{1}{c}{BLEU-1} & \multicolumn{1}{c}{BLEU-2} & \multicolumn{1}{c}{BLEU-3} & \multicolumn{1}{c}{BLEU-4} & \multicolumn{1}{c|}{ROUGE} & \multicolumn{1}{c}{BLEU-1} & \multicolumn{1}{c}{BLEU-2} & \multicolumn{1}{c}{BLEU-3} & \multicolumn{1}{c}{BLEU-4} & \multicolumn{1}{c}{ROUGE}  \\
\midrule
0.5                    & 35.75                      & 22.49                      & 15.84                      & 12.27                      & 35.94                      & 34.28                      & 21.05                      & 14.89                      & 11.53                      & 34.94                      \\
0.7                    & 37.23                      & 24.65                      & 18.09                      & 14.29                      & 38.39                      & 37.67                      & \textbf{25.43}             & \textbf{19.17}             & \textbf{15.38}             & 39.05                      \\
1.1                    & 36.24                      & 23.31                      & 16.94                      & 13.27                      & 36.55                      & 36.49                      & 23.33                      & 16.88                      & 13.13                      & 36.84                      \\
1.5                    & \textbf{38.61}             & \textbf{25.66}             & \textbf{18.75}             & \textbf{14.71}             & \textbf{38.91}             & \textbf{38.28}             & 24.97                      & 18.13                      & 14.08                      & \textbf{39.07}             \\
0.7 + 1.0 + 1.5         & 37.60                      & 23.64                      & 16.56                      & 12.65                      & 36.89                      & 36.67                      & 23.1                       & 16.66                      & 13.02                      & 36.92                      \\
\bottomrule
\end{tabular}
}

    \label{tab:ptt}
\end{table*}

\subsection{Experimental Setup}
\label{sec:experimental_setup}

\textbf{Dictionary} -  We collect our isolated dictionary for \ac{dgs}, from a range of sources, such as \cite{dgscorpus_3}. In total, we collect a \ac{dgs} sign vocabulary of 7,206 signs to experiment with. Given the \ac{ph14t} dataset as a gloss vocabulary of 1,066.  We are able to cover the vast majority of the vocabulary. However, we note the lack of specific place names.

\textbf{Dataset} -  To test our approach, we utilize the \ac{ph14t} dataset \cite{camgoz2018neural}. The dataset consists of 8,257 signed sentences in \ac{dgs}, each labelled at the gloss level and translated into German. The dataset is divided into training, development, and test sets, at a ratio of 80:10:10, respectively.

\textbf{Skeleton Keypoint} -  The skeleton keypoint representation comes from MediaPipe \cite{lugaresi2019mediapipe} and consists of 61 keypoints. 21 for each hand, 9 for the body, and 10 for the face. The 61 2D keypoints are uplifted to 3D using the method previously described \cite{10193629}. From the 3D skeleton, we solve for joint angles, which correspond to 104 angles. Each joint node in the skeleton contains between one to three degrees of freedom. We extract this representation for both the dictionary and the dataset. The dataset extraction is used to drive the signGAN model and train the \ac{ptt} model.

\textbf{Evaluation} - We evaluate the performance of our models using the \ac{bleu} \cite{papineni2002bleu} and \ac{rouge} score \cite{lin2004rouge}, which is a widely used metric for evaluating machine translation systems. Both scores measure the similarity between the generated translations and the reference translations.  \ac{bleu} score breaks down the translation into its word-level n-grams and calculates the precision of each, thus, we report \ac{bleu}-1 to 4 scores. We report the same metrics for both the \ac{ptt} and \ac{vtt} tasks.

\textbf{SignGAN} -  The generator network is trained for 68,000 iterations on a single signer. The generator network has 170,363,715 parameters, while the discriminator network has 5,535,618 parameters. The model is trained to output video at a resolution of 1080p. However, in line with the original \ac{ph14t} dataset, the videos are post-processed to a resolution of $256\times 256$ at 25 FPS. 

\textbf{SignSplat} - The Gaussian splatting appearance was trained on around 800 diverse frames with six camera views. The signer is rendered at 25 FPS at a resolution of $256\times 256$. For SMPL-X reconstruction, the complete optimisation takes approximately 20–40 seconds per short video (100–300 frames) on a NVIDIA GeForce RTX 3090.

\textbf{\ac{ptt} Model} -  For this, we utilise the publicly available ``Sign Language Transformers'' \cite{camgoz2020sign}, the same as \cite{walsh2024data, walsh2024sign, saunders2021signing, huang2021towards, saunders2020adversarial, saunders2021continuous, saunders2020progressive}.  We subsample the skeleton sequence to 12 frames per second for computational efficiency, and we normalise the skeleton such that the neck is set on the origin and the body is fixed on the \(xy\)-plane.  The model is trained for 200 epochs with a batch size of 256 and a learning rate of \(10^{-3}\). We construct the encoder and decoder to be symmetrical, both containing 8 heads and 3 layers, with an embedding dimension and feedforward size of 512. 
During training, dropout was utilized with a probability of 0.1, and the optimum beam size and alpha were found to be 3 and -1, respectively.
This model employs a reduced on-plateau scheduler with a patience of 5 and a decrease factor of 0.8.

\textbf{\ac{vtt} Model} -  For this we utilize the publicly available ``GF-SLT'' \cite{zhou2023gloss}. In line with the original paper, we perform 80 epochs of \ac{vlp}.  Here, the model employs a cosine scheduler with a warm-up of \(10^{-6}\) and a learning rate of \(5^{-3}\).  Following the \ac{vlp} phase, we perform an additional 200 epochs on the translation task, translating from \ac{vtt}. Employing the same learning rate scheduler, with an initial rate of \(10^{-2}\).  The text encoder is initialised from a multilingual BART model \cite{liu-etal-2020-multilingual-denoising}, which consists of 12 layers and is pre-trained on 25 languages.  Whereas the visual encoder consists of multiple layers of 2D and 1D convolutions, with ReLU and max-pooling.

\subsection{Quantitative Results}

\subsubsection{Pose-to-Text}
\label{sec:p2t}

\Cref{tab:ptt}.a shows the results of the \ac{ptt} task. We observe that the \ac{slp} techniques significantly improve the performance of the \ac{slt} model. Utilising the skeleton-based augmentation for pre-training outperforms the baseline model by a large margin, achieving an \ac{bleu}-1 score of 37.86 compared to 32.41 for the baseline. Higher n-grams also show similar results, with a \ac{bleu}-4 score increase of 2.36 on the test set.  This suggests that pre-training the model using different lexical variants not only allows the model to better recognise individual signs, but also improves the model's ability to understand the grammatical order in which they occur, as indicated by the increased \ac{bleu}-4 score.

 Row 2 of \Cref{tab:ptt}.a shows the results of training the model solely on the stitch data and then testing it on the original \ac{ph14t} test set.  Given the model has never seen real continuous sign language data, the results are impressive and indicate the stitch sequences contain features that are in line with the original.  We find that jointly training on both the real and the stitched data is detrimental to the model's performance on most metrics. Providing only marginal improvement for \ac{bleu}-3 and 4 scores on the dev set. Overall, the best performance is achieved with a pre-training phase followed by fine-tuning on the real data. 
 It provides at least 15\% improvement for all metrics, compared to the model trained only on real data.
 Therefore, the following two tables, \Cref{tab:ptt}.b and c, employ this pre-training strategy.

\begin{table*}[htbp!]
\centering
\caption[SignSplat and SignGAN Data Augmentation Results]{GFSLT Translation Metrics on the RWTH-PHOENIX-Weather-2014\textbf{T} DEV and TEST Sets}

\resizebox{0.95\linewidth}{!}{%
\begin{tabular}{l|ccccc|ccccc}
\toprule
                                    & \multicolumn{5}{c|}{TEST SET}                                                       & \multicolumn{5}{c}{DEV SET}                                                         \\
\multicolumn{1}{c|}{Training Data:} & BLEU-1         & BLEU-2         & BLEU-3         & BLEU-4         & ROUGE          & BLEU-1         & BLEU-2         & BLEU-3         & BLEU-4          & ROUGE           \\
\midrule
PHIX                                & 40.59          & 30.52          & 23.83          & 19.53           & 40.27          & 40.51          & 30.37          & 24.26          & 20.15          & 41.06          \\
PHIX (VLP PreTrain)                 & 43.06 & 32.68 & 25.88 & 21.35  & 42.67 & 43.39 & 32.94 & 26.24 & 21.65 & 43.62 \\
\hline
GAN                          & 2.41           & 0.95           & 0.38           & 0.19            & 4.69           & 2.71           & 1.00           & 0.42           & 0.19           & 5.19           \\
PHIX (Gan PreTrain)           & 41.46          & 31.20          & 24.65          & 20.28           & 41.63          & 41.93          & 31.68          & 25.17          & 20.85          & 42.37          \\
PHIX + GAN                   & 43.12          & 32.15          & 25.26          & 20.77           & 42.01          & \textbf{44.07} & \textbf{33.24} & 26.42          & 21.84          & \textbf{43.89}          \\
\hline
Splat                        & 6.51           & 2.20           & 0.83           & 0.40            & 5.30           & 7.13           & 2.55           & 1.15           & 0.62           & 5.85           \\
PHIX (Splat PreTrain)        & 41.15          & 30.79          & 23.90          & 19.47           & 40.14          & 42.50          & 31.92          & 25.38          & 20.97          & 42.18          \\
PHIX + Splat                 & \textbf{43.31} & \textbf{33.02} & \textbf{26.30} & \textbf{21.79} & \textbf{43.02} & 43.29 & 33.02 & \textbf{26.48} & \textbf{22.00} & 43.58 \\
\bottomrule
\end{tabular}
}

\label{tab:vtt}
\end{table*}

\begin{figure*}
    \centering
    \includegraphics[width=\linewidth, trim=100 0 100 0, clip]{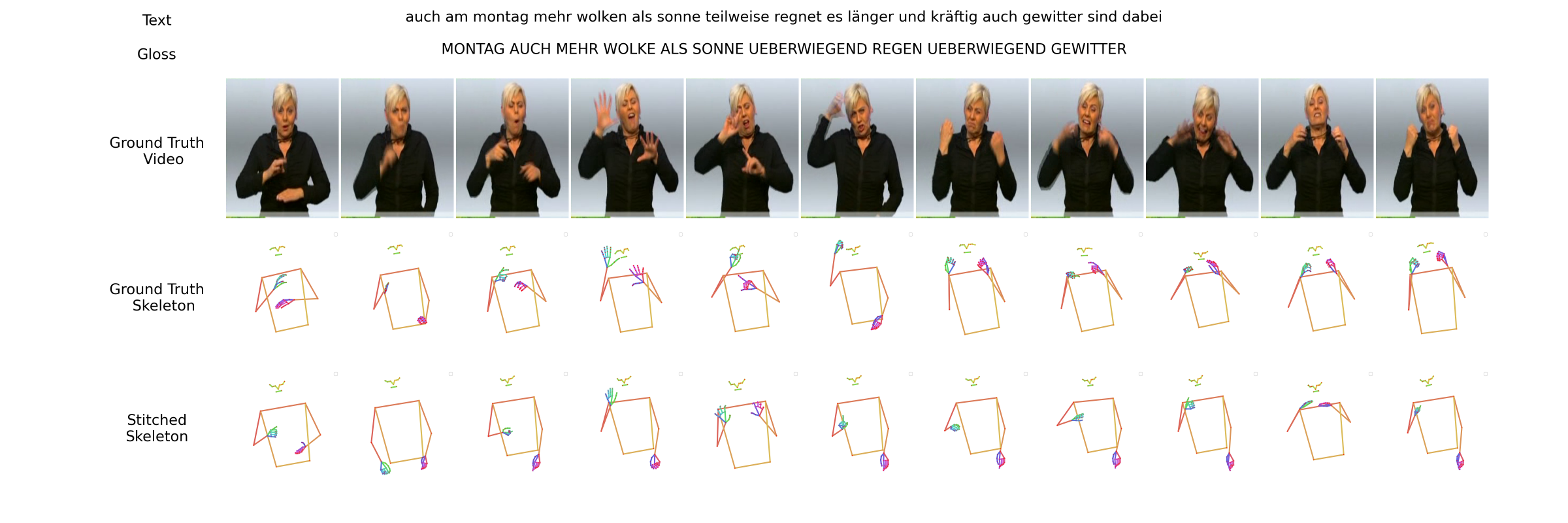}
    \caption{An example from the \ac{ph14t} dataset, showing top to bottom: spoken language, gloss, original video, extracted skeleton, and Stitched sequence.}
    \label{fig:skeleton_plots}
\end{figure*}

In \Cref{tab:ptt}.b we experiment with randomly permuting the order of N sequential signs in the pre-training data. We find that this has a marginal detrimental effect on the model's \ac{bleu}-1 score, decreasing it by up to 4\%. However, permuting up to 3 glosses in a sequence gives marginal improvements in the higher N-grams \ac{bleu} scores, indicating that creating small variations in the grammatical ordering makes the model more robust. However, permuting more than 3 glosses in a sequence results in a drop in performance. This suggests that the model is sensitive to the order of signs and that the grammatical structure of the sentences is important for achieving good translation results.

 Our final skeleton pose-based experiment investigates if creating variations in the signing speed in the pre-training data affects the performance on downstream tasks. As can be seen in \Cref{tab:ptt}.c  we determine that the best performance on the test set comes from increasing the speed by 1.5 times.  This is achieved by resampling the sequence to the desired number of frames. Possibly, reducing the total number of frames increased the difficulty in the pre-training data, allowing for the best performance on the Test set.  However, on the Dev set, we find that reducing the speed results in better \ac{bleu}-2 to 4 scores. Given only a 70\% sign overlap in the dev and test split, we suspect speed augmentation might introduce bias for some signs.

\subsubsection{Video-to-Text}
\label{sec:v2t}

\Cref{tab:vtt} shows the results of the \ac{vtt} task. We observe mixed results with the appearance-based augmentations.  In line with the original paper, we find that applying \ac{vlp} improves the baseline model's performance.  However, the best performance comes from including additional data created using the SignSplat approach. Increasing the \ac{bleu}-4 score on both the test and dev splits by 0.49 and 0.68, respectively. Surprisingly, given that the \ac{ph14t} dataset contains only 9 signers and most models overfit to the appearance of these individuals. We hypothesise that greater performance gains can be achieved by including more appearances rendered from multiple viewpoints.

 We achieve inconsistent performance using the data generated with the SignGAN approach, even though the qualitative results show the model produces a realistic synthetic signer. Artifacts caused by noise in the generation process can degrade the quality of the data. Issues such as disconnected limbs, loss of detail between fingers and hands, and artefacts in the face are all possible causes.  On the Dev set, we find the best improvements in \ac{bleu}-1 and 2 score using this data, while it is detrimental on most other metrics.

\subsection{Qualitative Results}
\label{sec:qualitative_results}

\begin{figure*}
    \centering
    \includegraphics[width=\linewidth, trim=100 0 100 0, clip]{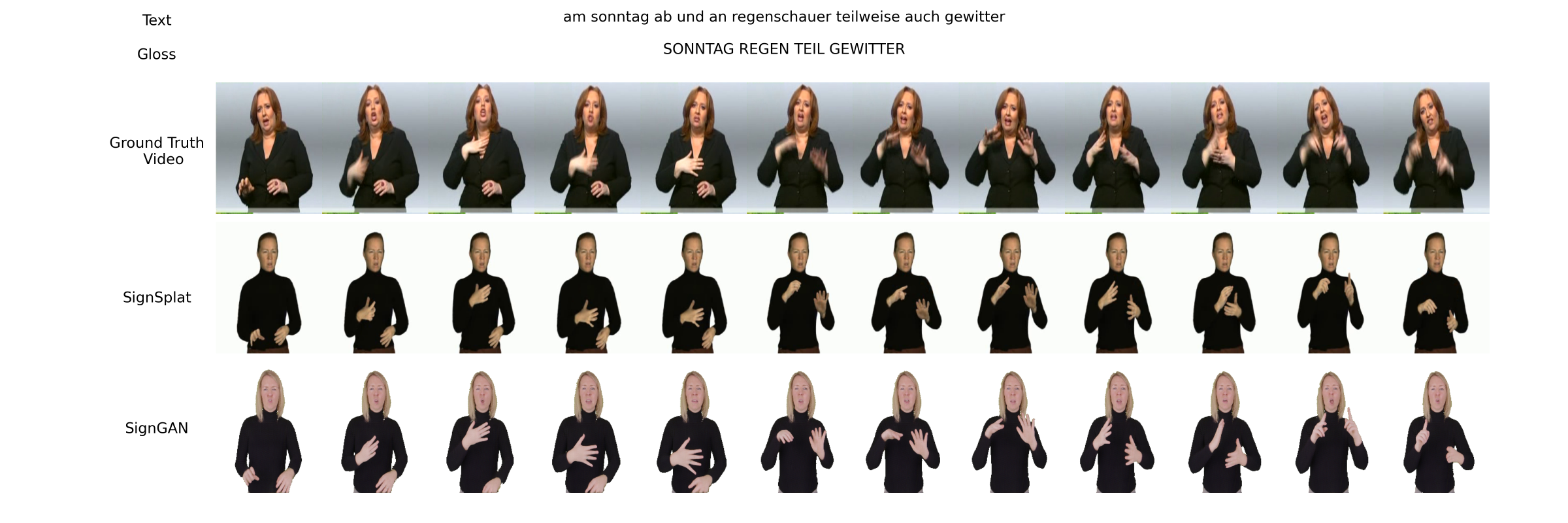}
    \caption{An example from the \ac{ph14t} dataset, showing top to bottom: spoken language, gloss of the original video, SignSplat Avatar, and SignGAN Avatar.}
    \label{fig:gan_plots}
\end{figure*}

\Cref{fig:skeleton_plots} and \Cref{fig:gan_plots} show examples of the generated data from the \ac{ph14t} dataset.  Both figures show the original video, the gloss, and the spoken language translation.  \Cref{fig:skeleton_plots} shows the extracted skeleton and the stitched sequence.  We note that the two sequences share many features like handshape and locations. However, the temporal alignment between the two differs, mostly due to our approximations, as the timing information is not available in the metadata.

\Cref{fig:gan_plots} shows the visual augmentation from both SignSplat and SignGAN. Both approaches can faithfully capture the motion and maintain a realistic appearance, although with slight changes in camera angle compared to the ground truth.

\begin{figure}
    \centering
    \includegraphics[width=0.8\linewidth, trim=0 0 0 0, clip]{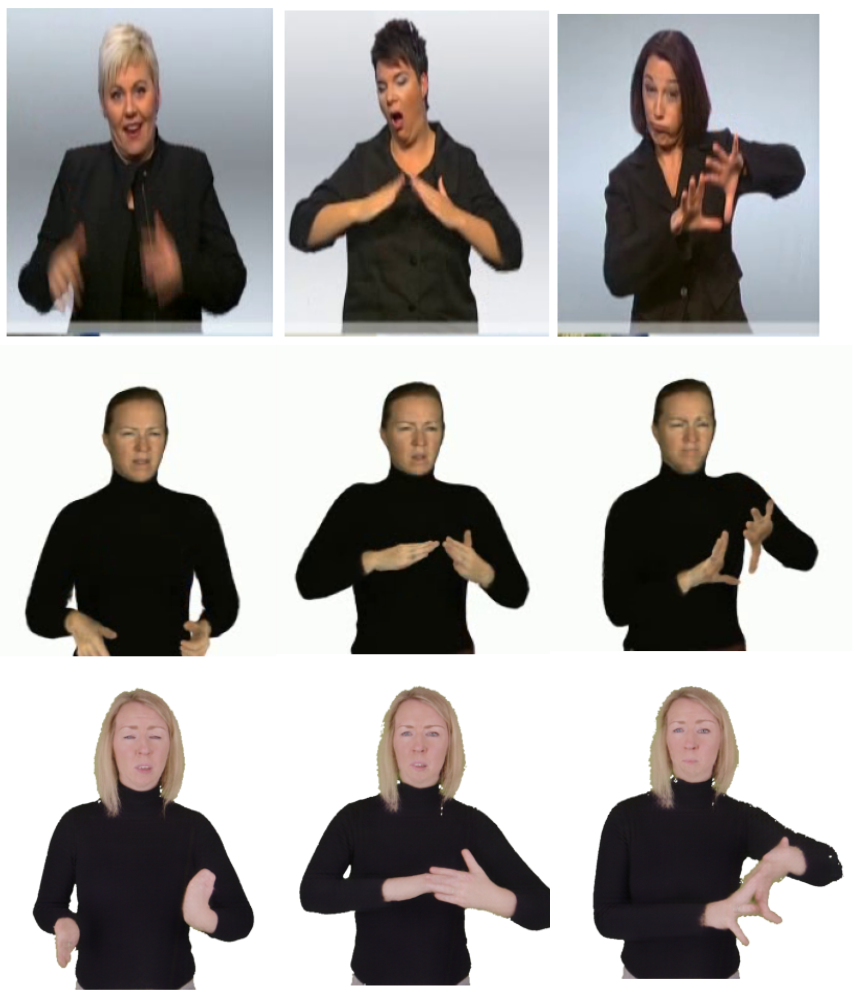}
    \caption{Examples from the \ac{ph14t} dataset, showing top to bottom: the original video, SignSplat Avatar, and SignGAN Avatar. The bottom row shows issues with blurring, hand merging and artefacts in the face.}
    \label{fig:error_plots}
\end{figure}

The SignSplat model can produce a more realistic signer, with fewer artefacts and a more natural appearance.  The SignGAN model, while able to produce realistic video sequences, suffers from artefacts and noise in the data, which can detract from the overall quality of the generated video.  This is likely due to the limitations of the GAN architecture and the training data used. As shown in \Cref{fig:error_plots}, some of the artefacts include loss of detail between fingers and hands, and artefacts in the face.

\section{Conclusion}
\label{sec:conclusion}
In this paper, we propose leveraging three different \ac{slp} techniques to augment and generate synthetic sign language data.  We discovered that stitching together isolated signs to create synthetic skeleton sequences allowed us to pretrain SLT models, leading to significant performance gains. Further improvements across all metrics can be achieved by introducing temporal variations in these sequences. For high N-gram scores (specifically BLEU-3 to BLEU-4), creating grammatical variations also proves beneficial.
 

We tested two different approaches for generating visual augmentations that were able to transfer the motion of the original dataset to an entirely new appearance. Qualitatively, we found that the GAN was able to produce realistic video sequences, but was prone to unrealistic artefacts caused by noise in the data. This method proved effective on a subset of metrics. The Gaussian Splatting approach was able to produce realistic sequences with fewer artefacts, given that the approach is constrained by the underlying human mesh. As a result, we demonstrated the benefits of introducing visual augmentations.

 Given the low-resource nature of sign language translation datasets, we suggest this is a promising direction for future research. These simple approaches could be considered a baseline for many more experiments, which we believe these results indicate could generate much greater improvements in performance. For instance, combining the visual augmentations with the novel skeleton motion to generate entirely new video sequences rendered from multiple viewpoints. Furthermore, we believe the power of LLMs is yet to be leveraged here to generate novel sentences.



\begin{acks}
This work was supported by the SNSF project `SMILE II' (CRSII5 193686), the Innosuisse IICT Flagship (PFFS-21-47), EPSRC grant APP24554 (SignGPT-EP/Z535370/1), Google DeepMind and through funding from Google.org via the AI for Global Goals scheme. This work reflects only the author’s views and the funders are not responsible for any use that may be made of the information it contains.
\end{acks}

\newpage
\bibliographystyle{ACM-Reference-Format}
\bibliography{main}


\end{document}